%% file: Deep-FPF_final.tex
\documentclass[letterpaper, 10 pt, conference]{ieeeconf}  

\IEEEoverridecommandlockouts 

\usepackage{afterpage}
\usepackage{epsfig} 
\usepackage{times} 
\usepackage{amsmath} 
\usepackage{bbm}
\usepackage{amssymb}  
\usepackage{amsfonts}
\usepackage{mathptmx} 
\usepackage{tabularx}
\usepackage{epstopdf}
\usepackage{float}
\usepackage{epic,color}
\usepackage{mathrsfs}
\usepackage{dsfont,MnSymbol}
\usepackage{multirow} 
\usepackage[T1]{fontenc}
\usepackage{tikz}
\usepackage{textcomp}

\usepackage{algorithmic, algorithm}

\usepackage{graphicx,color,fancyhdr}
\usepackage{subcaption}

\newcommand{\calF}{\mathcal{F}}

\newcommand\copyrighttext{%
  \footnotesize \textcopyright 2020 IEEE. Personal use of this material is permitted.
  Permission from IEEE must be obtained for all other uses, in any current or future
  media, including reprinting/republishing this material for advertising or promotional
  purposes, creating new collective works, for resale or redistribution to servers or
  lists, or reuse of any copyrighted component of this work in other works.}
\newcommand\copyrightnotice{%
\begin{tikzpicture}[remember picture,overlay]
\node[anchor=south,yshift=10pt] at (current page.south) {\fbox{\parbox{\dimexpr\textwidth-\fboxsep-\fboxrule\relax}{\copyrighttext}}};
\end{tikzpicture}%
}

\newcounter{rmnum}
\newenvironment{romannum}{\begin{list}{{\upshape (\roman{rmnum})}}{\usecounter{rmnum}
\setlength{\leftmargin}{22pt}
\setlength{\rightmargin}{8pt}
\setlength{\itemsep}{2pt}
\setlength{\itemindent}{-1pt}
}}{\end{list}}

\newcounter{anum}

\def\IEEEQEDclosed{\mbox{\rule[0pt]{1.3ex}{1.3ex}}}

\def\qed{\ifmmode\IEEEQEDclosed\else{\unskip\nobreak\hfil
		\penalty50\hskip1em\null\nobreak\hfil\IEEEQEDclosed
		\parfillskip=0pt\finalhyphendemerits=0\endgraf}\fi}

\input{_definitions}

\def\clZ{{\cal Z}}

\newlength{\noteWidth}
\long\def\notes#1{\ifinner
	{\tiny #1}
	\else
	\marginpar{\parbox[t]{\noteWidth}{\raggedright\tiny #1}}
	\fi}

\title{\LARGE \bf Deep FPF: Gain function approximation  in high-dimensional setting}

\author{S. Yagiz Olmez, Amirhossein Taghvaei and Prashant G. Mehta
	\thanks{Financial support from the NSF grant 1761622 and the ARO grant W911NF1810334 is gratefully acknowledged.}
	\thanks{S.~Y. Olmez and P.~G.~Mehta are with the Coordinated
		Science Laboratory and the Department of Mechanical Science and
		Engineering at the University of Illinois at Urbana-Champaign
		(UIUC); A. Taghvaei is with the Department of Mechanical and Aerospace Engineering University of California Irvine; Corresponding email: mehtapg@illinois.edu.}
}

\begin{document}

\maketitle
\copyrightnotice

\thispagestyle{empty}
\pagestyle{empty}

\begin{abstract}
%
%
In this paper, we present a novel approach to approximate the gain
function of the feedback particle filter (FPF). The exact gain
function is the solution of a Poisson equation involving a
probability-weighted Laplacian. The numerical problem is to
approximate the exact gain function using only finitely many particles
sampled from the probability distribution.   

Inspired by the recent success of the deep learning methods, we
represent the gain function as a gradient of the output of a neural
network.  Thereupon considering a certain variational formulation of
the Poisson equation, an optimization problem is posed for learning
the weights of the neural network.  A stochastic gradient algorithm is
described for this purpose.    

The proposed approach has two significant properties/advantages: (i) The
stochastic optimization algorithm allows one to process, in parallel, only a batch
of samples (particles) ensuring good scaling properties with the
number of particles; 
(ii) The remarkable representation power of neural networks means that
the algorithm is potentially applicable and useful to solve high-dimensional
problems. We numerically establish these two properties and provide
extensive comparison to the existing approaches.

\end{abstract}

\section{Introduction}
\label{sec:intro}
This research is motivated by the following questions: What is the
principled approach to apply the deep
learning methodology to the stochastic filtering problem? 
Can the well known curse of dimensionality in these problems be
mitigated by learning certain geometric structures using neural
networks?

In this paper, we present a principled approach to address these
questions, based on the use of the feedback particle filter (FPF) methodology.
The FPF methodology is applicable to the continuous-time stochastic
filtering problem modeled by the following nonlinear stochastic differential
equations (sde)~\cite{xiong2008}:
\begin{subequations}
\begin{align}\label{eq:filtering_states}
\ud X_t &= a(X_t)dt + \sigma(X_t) \ud B_t, \\\label{eq:filtering_obs}
\ud Z_t &= h(X_t)dt + \ud W_t,
\end{align}
\end{subequations}
where  $X_t \in \Re^d$ is the state of a hidden Markov process at time
$t$,  $Z_t \in \Re$ is the observation process, $\{B_t\}_{t\geq 0}$
and $\{W_t\}_{t\geq 0}$ are two mutually independent Wiener
processes, and the functions $a(\cdot)$, $\sigma(\cdot)$, and
$h(\cdot)$ are assumed to be continuously differentiable. 
The objective of the filtering problem is to compute the {\it posterior
  distribution}, i.e., the conditional distribution of $X_t$ given the
history of observations $\clZ_t:=\sigma(Z_t;t\in[0,t])$. 

The most commonly used approach to approximate the solution of the nonlinear filtering problem are sequentially importance sampling and resampling particle filters~\cite{gordon93,doucet09}. However, these approaches are known to perform poorly in high-dimensional setting (when $d$ is large), an issue known as curse of dimensionality~\cite{bickel2008sharp,bengtsson08,snyder2008obstacles,rebeschini2015can}. Feedback particle filter  
is an alternative algorithm that does not involve the importance
sampling and resampling steps~\cite{taoyang_TAC12,yang2016}. FPF
algorithm comprises of a system of $N$ stochastic processes
$\{X^i_t;1\leq i\leq N,\;t\geq 0\}$, referred to as particles, driven by a control law designed such that the empirical distribution of the particles  approximates the posterior distribution of the filter.  In particular, the evolution of $i$-th particles is governed by the following sde:  
\begin{equation}\label{eq:mean_field}
\ud X_t^i = a(X_t^i)\ud t+\sigma(X_t^i) \ud B_t^i+ \underbrace{{\sf K}_t(X_t^i)\circ(\ud Z_t-\frac{h(X_t^i)-\hat{h}_t}{2}\ud t)}_{\text{feedback control law }},
\end{equation}
where ${\sf K}_t(\cdot)$ is the so-called {\it gain function},
$\hat{h}_t := \Expect[h(X^i_t)|\clZ_t]$, $\{B^i_t\}_{t\geq 0}$ is an independent
copy  of $\{B_t\}_{t\geq 0}$, and $\circ$ indicates Stratonovich integration.  The
gain function 
$${\sf K}_t(\cdot)=\nabla \phi_t(\cdot ),
$$ 
where the function $\phi_t(\cdot)$ solves the probability weighted Poisson equation  
	\begin{align}\label{eq:Poisson}
	&-\frac{1}{\rho_t(x)}\nabla \cdot(\rho_t(x) \nabla \phi_t(x)) = h(x) -\hat{h}_t,
	\end{align}
	where $\rho_t(\cdot)$ is the probability density function for
        $X^i_t$, and $\nabla \cdot$ denotes the divergence operator. 
Although FPF does not involve importance sampling and resampling
steps, its implementation is computationally challenging because of
the numerical problem of gain function approximation.  It is noted
that in a numerical simulation the density $\rho_t(\cdot)$ is not
explicitly available.  The Poisson equation must be solved by using
{\em only} the particles $\{X^i_t\}_{i=1}^N$ which -- for the purposes of
analysis and algorithm development -- are assumed to be independent
samples drawn from $\rho_t$.  Although the considerations of this
paper are motivated by the FPF algorithm for the continuous-time
filtering problem, the gain function approximation is also the central
problem for the discrete-time FPF model~\cite{yang2014continuous} and also for the
particle flow algorithm~\cite{daum10,daum2017generalized}.




In literature, there are two approaches to the numerical problem of
gain function approximation: the Galerkin approach~\cite{yang2016} and
the diffusion map-based approach~\cite{Amir_CDC2016,Amir2019SIAM}.  As
illustrated with examples in Sec.~\ref{subsec:scaling_with_d}, these existing
approaches do not scale well with the problem dimension.  The Galerkin
procedure requires selection of basis functions which becomes unwieldy
as the problem dimension becomes large.  The diffusion map-based
algorithm can require an exponentially large number of particles, with
respect to the problem dimension, in order to maintain the same amount
of error.

In this paper, we present a novel deep learning-inspired approach to
mitigate some of the existing limitations. Our proposed approach is
based on the variational formulation of the Poisson equation. In
particular, the solution of the Poisson equation~\eqref{eq:Poisson}
represents the minimizer of the following variational problem:
\begin{equation}\label{eq:var-problem}
\min_{f\in H^1_0(\rho_t)}\,\int_{\Re^d}\left( \half |\nabla f(x)|^2 - (h(x)-\hat{h}_t) f(x)\right)\rho_t(x)\ud x,
\end{equation}
where $H^1_0(\rho_t)$ is the Hilbert space of  square integrable (with
respect to $\rho_t$)
functions whose derivative (defined in weak sense) is also square
integrable.  We restrict the function class $H^1_0(\rho_t)$ to a
parametric family of functions $\calF_\Theta$ represented by
feedforward neural networks.   On $\calF_\Theta$, the variational problem
is empirically approximated in terms of the particles:
\begin{equation}\label{eq:opt-particles}
\min_{f \in \calF_\Theta}\, \frac{1}{N}\sum_{i=1}^N \left( \half
  |\nabla f(X^i_t)|^2 - (h(X^i_t)-\hat{h}^{(N)}_t)f(X^i_t) \right),
\end{equation}
where $\hat{h}^{(N)}_t = N^{-1}\sum_{i=1}^N h(X^i_t)$. 
A stochastic gradient descent algorithm is proposed to learn the
parameters of the network by solving the empirical optimization
problem~\eqref{eq:opt-particles}.  Finally, the gain function is
approximated as ${\sf K}_{NN}(\cdot) = \nabla f^*(\cdot)$ where
$f^*(\cdot)$ is the output of the optimized neural network.

Our proposed approach has two significant advantages: 
\begin{romannum}
\item  The stochastic optimization algorithm allows one to process
  only a batch of particles with size $M\ll N$. Moreover, these
  computations can be done in parallel for each particle. This is a
  significant improvement over, e.g., the diffusion map-based
  algorithm where the computations scale with $O(N^2)$. 
\item The expressive power of neural network architecture allows the
  algorithm to potentially scale better to high-dimensional settings
  with complicated probability distributions. The diffusion-map based
  algorithm does not scale well because it is based on a Gaussian
  kernel which becomes progressively singular in high-dimensional setting.   
\end{romannum}

In recent years, there has been a growing interest in solving partial
differential equations (PDEs) using the deep learning
methodology~\cite{deepxde,dgm,deepritz}.  Related to the construction
described in our paper,~\cite{deepritz} introduces an approach based
on a variational formulation of a PDE.

The outline of the remainder of this paper is as follows: The
consistency of the variational formulation and its stability analysis
appears in Sec.~\ref{sec:variational}. The proposed numerical
procedure and neural network architecture appears in
Sec.~\ref{sec:emprical}. Numerical experiments and comparison to
existing approaches appear in Sec.~\ref{sec:numerical}. The
application to filtering problem appears in
Sec.~\ref{sec:filtering_application} and the conclusions in Sec.~\ref{sec:conclusion}.


\section{Variational formulation}
\label{sec:variational}
Let $J(\cdot)$ denote the objective functional of the variational problem~\eqref{eq:var-problem}
\begin{equation}\label{eq:J-def}
J(f) := \int_{\Re^d}\left( \half |\nabla f(x)|^2 - (h(x)-\hat{h})
  f(x) \right) \rho(x)\ud x,
\end{equation} 
where we dropped the time index $t$ for clarity of presentation.  A
formal calculation shows that the weak form of the Poisson equation~\eqref{eq:weak-form} is the
first order optimality condition of~\eqref{eq:var-problem}: if $\phi_0$ is the minimizer of~\eqref{eq:var-problem}, then 
\begin{equation}
\begin{split}
0&=\left. \frac{\ud}{\ud \epsilon} J(\phi_0+\epsilon \psi) \right|_{\epsilon=0} \\
&= \int_{\Re^d} \left[ \langle \nabla \phi_0(x),\nabla \psi(x) \rangle - (h(x)-\hat{h}_t)  \psi(x) \right]   \rho(x)\ud x.
\end{split}
\end{equation} 
for all functions $\psi \in H^1_0(\rho)$, where $\langle \cdot,\cdot\rangle$ denotes the inner product on $\Re^d$. 
A more rigorous analysis requires additional assumptions on the
density $\rho$ and the function $h$:

\newP{Assumption A1} (i) The probability density $\rho$ satisfies the Poincar\'e inequality, i.e. 
\begin{equation}
\label{eq:poincare}
\int_{\Re^d} |f(x)-\hat{f}|^2\rho(x) \ud x \leq \int_{\Re^d} |\nabla f(x)|^2 \rho(x)\ud x,\quad \forall f \in H^1(\rho),
\end{equation}
where $\hat{f}= \int f(x) \rho(x) \ud x$.  (ii) The function $h$ is square integrable with respect to $\rho$, i.e. $h\in
L^2(\rho)$.  

\medskip

\begin{theorem}
	Under the assumption A1, the variational
        problem~\eqref{eq:var-problem} has a unique minimizer, denoted
        by $\phi_0$, and the minimizer solves the weak form of the
        Poisson equation~\eqref{eq:Poisson}:
	\begin{equation}\label{eq:weak-form}
	\int_{\Re^d} \langle \nabla \phi_0,\nabla \psi \rangle \rho \ud x = \int_{\Re^d} (h-\hat{h}) \psi \rho \ud x,\quad \forall \psi \in H^1(\rho).
	\end{equation}  
\end{theorem}
\medskip

\begin{proof}
	The proof is generalization of the arguments used to analyze the variational formulation of the classical Poisson equation, c.f.~\cite[Sec. 3.10]{Laugesen2015linear}. The key steps of the proof are (i) showing a lower-bound on the objective function~\eqref{eq:J-def} using the Poincar\'e inequality; (ii) construction of minimizing sequence $(f_n)$; (iii) weak convergence of the sequence to function $\phi_0$; (iv) and showing that $\phi_0$ is the minimizer by using lower-semicontinuity of $J(\cdot)$.  
\end{proof}
\medskip

In practice, any numerical procedure will necessarily yield an
approximation of the exact gain function.  The following
proposition characterizes the $L^2$-error in approximation.

\medskip


\begin{proposition}\label{lem:J-stability}
Consider the variational formulation~\eqref{eq:var-problem} with unique minimizer $\phi_0$.  Then,
\begin{equation}
J(\phi) =  J(\phi_0) + \half \|\nabla \phi  - \nabla  \phi_0\|_{L^2(\rho)}^2.
\end{equation}
\end{proposition}

\medskip

\begin{proof}
	The proof follows by decomposing $\|\nabla \phi  - \nabla  \phi_0\|_{L^2(\rho)}^2 $ and using  the following two identities: 
	\begin{equation}
	\begin{split}
 \int_{\Re^d} \langle \nabla \phi_0,\nabla \phi \rangle \rho \ud x &=  \int_{\Re^d}  (h-\hat{h})\phi \rho \ud x, \\
J(\phi_0) &=  -\frac{1}{2}\int_{\Re^d} \|\nabla \phi_0\|^2 \rho \ud x.
	\end{split}
	\end{equation}
The identities follow from the optimality condition~\eqref{eq:weak-form} with $\psi=\phi$ and $\psi=\phi_0$ respectively. 
\end{proof}


%

\section{Proposed numerical approach}
\label{sec:emprical}

\subsection{Empirical approximation}
The empirical approximation of the objective function $J(\cdot)$
in~\eqref{eq:J-def} is defined as follows
\begin{equation}\label{eq:hatJN}
\hat{J}^{(N)}(f) := \frac{1}{N} \sum_{i=1}^N \half |\nabla f(X^i)|^2 - f(X^i)(h(X^i)-\hat{h}^{(N)}),
\end{equation} 
where $\{X^i\}_{i=1}^N$ are assumed to be independent samples
distributed according to $\rho$, and  $\hat{h}^{N} = \frac{1}{N}
\sum_{i=1}^N h(X^i)$. Minimizing the empirical
approximation~\eqref{eq:hatJN} over all functions is ill-posed: the
minimum is unbounded and minimizer does not exist.  This is because
the empirical probability distribution does not satisfy the Poincar\'e
inequality.
Hence, we restrict the function class for the optimization problem and
consider 
\begin{equation}\label{eq:empirical-opt}
\min_{f_\theta \in \calF_\Theta}~\hat{J}^{(N)}(f_\theta),
\end{equation}
where $\calF_\Theta$ is a parameterized class of functions. A function
in the class $\calF_\Theta$ is denoted by $f_\theta(x)$ or $f(x;\theta)$ where $\theta \in \Theta$ is the parameter, and $\Theta$ is the parameter set. For example 
\begin{enumerate}
	\item $\calF_\Theta = \{ \sum_{j=1}^m \theta_i \psi_j; ~
          \psi_j \in H^1_0, \theta_j \in \mathbb{R} \text{ for }
          j=1,\ldots,m\}$ is a linear combination of selected basis
          functions. This linear parametrization leads to the Galerkin algorithm. 
	\item $\calF_\Theta$ is represented with  neural networks and
          the parameters are the weights in the network.   
\end{enumerate}

In this paper, we follow the neural network representation and propose the following neural network architecture.

\subsection{Neural network architecture} The output $f(x;\theta)$ of
the network is defined by the following feed-forward network:
\begin{equation}
\begin{split}
f(x;\theta) &= h_L,\\
h_{l+1} &= \sigma_l(W_lh_l+b_l+A_lx),\quad l=0,1,\ldots,L-1,
\end{split}
\end{equation}  
where $x$ is the input, $W_l$ and $A_l$ are weight matrices (with the
convention that $W_0=0$), $b_l$ is the bias term,  $\sigma_l$ is the
activation function at layer $l$, and $L$ is the number of layers or
depth of network.

The choice of activation function has an effect on the representation power of the neural network. We consider the following selections for activation functions: 
\begin{romannum}
	\item The first activation function $\sigma_1(x)=\max(x,\alpha x)^2$ is the square of the leaky ReLU function, where $\alpha<1$ is the leak parameter.
	\item The activation functions $\sigma_l(x)=\max(x,\alpha x)$ is the leaky ReLU for $l=2,\ldots,L-2$. 
	\item The last activation function $\sigma_{L-1}(x)=x$ is identity. 
\end{romannum}
The choice for leaky ReLU compared with ReLU ensures that the gradient
does not vanish. This has been found to be helpful for the
optimization procedure. The choice for square ReLU for
$\sigma_1(\cdot)$ ensures that the output is piecewise
quadratic. Therefore, the gradient of the output, which represents the
gain function, is piecewise affine. Without the square, the gradient
of output is piecewise constant. Numerically, we observed that
piecewise affine approximations are better when compared to the
piecewise constant approximations. Finally, the identity map is used
for the last layer is to ensure that the neural network can easily
represent affine maps.

 \subsection{Optimization procedure}
%

The optimization problem~\eqref{eq:empirical-opt} is solved using the
Adam optimizer~\cite{Adam}. At each iteration, a batch of particles is
selected randomly  and used to update the weights of the network. The
algorithm is implemented using the existing tensor-flow libraries and
modules. A summary of the proposed numerical procedure is presented as
Algorithm~1.

\begin{algorithm}
\caption{Numerical procedure to solve~\eqref{eq:empirical-opt}}
\label{alg:DeepGain}
	\begin{algorithmic}
		\STATE {\bfseries Input:} particles $\{X^i\}_{i=1}^N$, observation values $\{h(X^i)\}_{i=1}^N$ Batch size $M$, Total iterations $T$ 
		\FOR{$t=1,\ldots, T$}
		\STATE Sample batch $\{X_i\}_{i=1}^{M} $ from $\{X^i\}_{i=1}^N$ 
		\STATE	Update $\theta$ to minimize~\eqref{eq:empirical-opt} using Adam method
		\ENDFOR
	\end{algorithmic}
\end{algorithm}

\section{Numerical experiments}
\label{sec:numerical}
For the following reported experiments, the algorithm parameters are set as in Table \ref{table:params}, unless stated otherwise.
%
%
%

\begin{table}[h]
	\caption{Algorithm parameters}
	\label{table:params}
	\centering
	\begin{tabular}{|c|c|c|} \hline
		description & notation & value \\ \hline
		number of layers & $L$ & $4$ \\
		number of neurons & $m$ & $32$ \\
		leak parameter for ReLU& $\alpha$ & $0.3$ \\
		batch size & $M$ & $10$  \\
		sample size & $N$ & $100$ \\
		number of iterations & $T$ & $10^4$\\
		learning rate & $\eta$ & $10^{-4}$\\
		adam parameters & $\beta_1,\beta_2$ & $0.9,0.999$ 
		\\\hline
	\end{tabular}
\end{table}

\begin{figure*}[t]
	\centering
	\begin{tabular}{ccc}
		\begin{subfigure}[b]{0.32\linewidth}
			\includegraphics[width=0.95\linewidth]{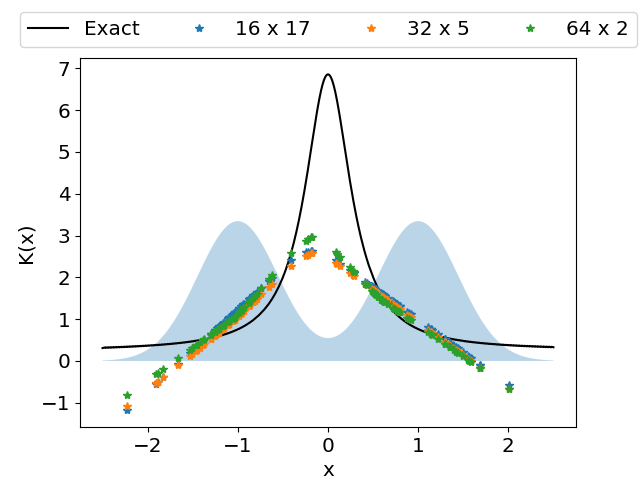}
			\caption{Deep FPF as in Table~\ref{alg:DeepGain}}
		\end{subfigure}
		&
		\begin{subfigure}[b]{0.32\linewidth}
		\includegraphics[width=0.95\linewidth]{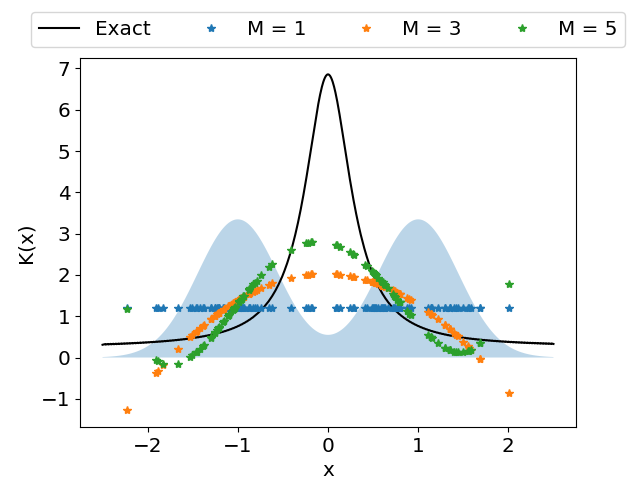}
		\caption{Galerkin with polynomial basis.}
	\end{subfigure}
	&
	\begin{subfigure}[b]{0.32\linewidth}
		\includegraphics[width=0.95\linewidth]{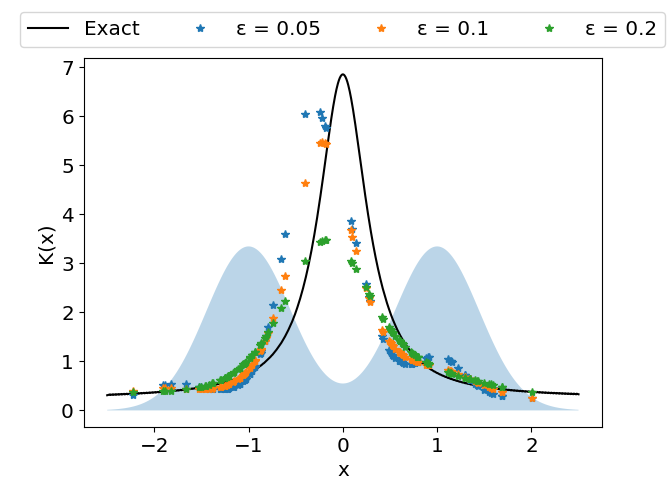}
		\caption{Diffusion Map Based (DM)}
	\end{subfigure}
	\end{tabular}
	\caption{Gain function approximation for bimodal example in Section~\ref{subsec:bimodal}.}
	\label{fig:bimodal}
\end{figure*}

\subsection{Illustration with bimodal distribution}
\label{subsec:bimodal}
The distribution $\rho$ is assumed to be bimodal distribution $\frac{1}{2}N(-1,\sigma^2) + \frac{1}{2}N(1,\sigma^2)$ where $\sigma^2=0.2$. The proposed numerical procedure in Table~\ref{alg:DeepGain} is implemented to approximate the gain function. The following three selection of architectures parameters are used 
\begin{itemize}
		\item[(i)] $2$ layers $64$ neurons each layer
		\item[(ii)] $5$ layers $32$ neurons each layer
	\item[(iii)] $17$ layers, $16$ neurons each layer
\end{itemize}
The three architectures involve same number of unknown weight parameters. 

The result is compared with the exact gain. The exact gain function admits a formula in the scalar case given by
\begin{equation}\label{eq:K_bimodal_exact}
{\sf K}(x) =  - \frac{1}{\rho(x)}\int_{-\infty}^x \rho(z)(h(z)-\hat{h})dz.
\end{equation}

The numerical  results are depicted in Figure~\ref{fig:bimodal}(a): All three architectures produce the same result. For comparison, the Galerkin and diffusion map-based methods are implemented for the bimodal example. The results are depicted in Figure~\ref{fig:bimodal}(b) and Figure~\ref{fig:bimodal}(c) respectively. The details of Galerkin algorithm  and diffusion map-based algorithm appear in \cite{yang2016} and \cite{Amir2019SIAM} respectively. The polynomial basis function is used for the Galerkin algorithm.

For this particular example, Deep FPF outperforms Galerkin approximation but does not perform as well as diffusion map based approximation. However, note that there is a lot of room for neural networks to be tuned to produce better result. This is subject of ongoing work. 

\subsection{Over-fitting with full batch optimization}
\begin{figure*}[t]
	\centering
	\begin{tabular}{cc}
	\begin{subfigure}[b]{0.4\linewidth}
		\includegraphics[width=\linewidth]{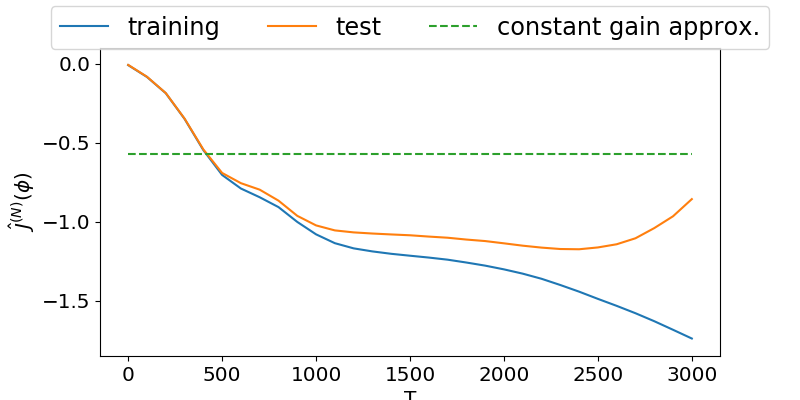}
		\caption{N = M = 100}
	\end{subfigure}
	&
	\begin{subfigure}[b]{0.4\linewidth}
		\includegraphics[width=\linewidth]{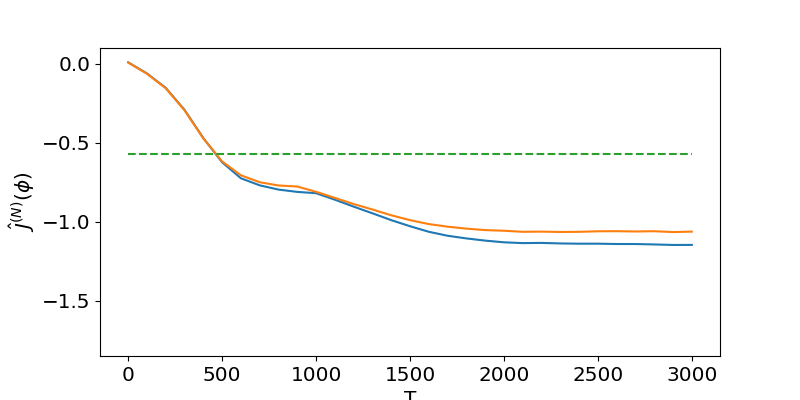}
		\caption{N = 100 and M = 10}
	\end{subfigure}
\end{tabular}
	\caption{The training and test error of the empirical objective function~\eqref{eq:hatJN} for application the numerical procedure~\ref{alg:DeepGain} on the bimodal example of Section~\ref{subsec:bimodal}.}
	\label{fig:mse_of}
\end{figure*}

Consider the application of the proposed procedure on the bimodal example in Section~\ref{subsec:bimodal} when the batch-size is equal to the number of samples, i.e. $M=N=100$. The value of the empirical objective function~\eqref{eq:hatJN} evaluated on the given samples $\{X^i\}_{i=1}^N$ (training loss), and evaluated on fresh independent samples $\{Y_i\}_{i=1}^{N_1}$ (test loss), as a function of iteration, are depicted in Figure~\ref{fig:mse_of}-(a). It is observed that after certain iteration, around $2000$, the test loss starts to increase while the training loss continues to decrease. This illustrates the over-fitting phenomenon. The gap between training loss and test loss is called generalization error. 

The over-fitting is avoided when the batch-size is smaller, $M=10$. The training loss and test loss are depicted in~\ref{fig:mse_of}-(b). It is observed that both training loss and test loss continue to decrease as the iteration number grows.  It is also observed that the generalization error remains small.   The reason is that random selection of batches at each iteration of the optimization algorithm introduces randomness that prevents over-fitting~\cite{bottou2010large}.   


\subsection{Scaling with dimension}
\label{subsec:scaling_with_d}

Consider the following probability density function
\begin{equation}
\rho(x) = \rho_b(x_1) \prod_{n=2}^d \rho_g (x_n),
\end{equation}
for $x = (x_1,x_2,...,x_d) \in \mathbb{R}^d$. $\rho_b$ is the probability density function for bimodal distribution introduced in Section~\ref{subsec:bimodal} and $\rho_g$ is the probability density function for $N(0,\sigma^2)$. Assume that the observation function is $h(x) = x_1$. Hence the exact gain function is given by
\begin{equation}
\tilde{ \sf K}_{\text{exact}}(x) = ({\sf K}_{\text{exact}}(x_1),0,...,0),
\end{equation}
where ${\sf K}_{exact}(x_1)$ is given by~\eqref{eq:K_bimodal_exact}.

Define the m.s.e error according to
\begin{equation}\label{eq:mse}
\text{m.s.e} = \frac{1}{N} \sum_{i=1}^N |{\sf K}_{\text{alg.}}(Y^i) - {\sf K}_{\text{exact}}(Y^i)|^2,
\end{equation}
where $\{Y^i\}_{i=1}^N$ are independent samples from $\rho$, and ${\sf K}(\cdot)$ is the approximate gain obtained from the proposed algorithm~\ref{alg:DeepGain}. The m.s.e is computed by averaging over $100$ simulations. The sample size $N=1000$.

\begin{figure*}[t]
	\centering
	\begin{tabular}{ccc}
	\begin{subfigure}[b]{0.3\linewidth}
		\includegraphics[width=\linewidth]{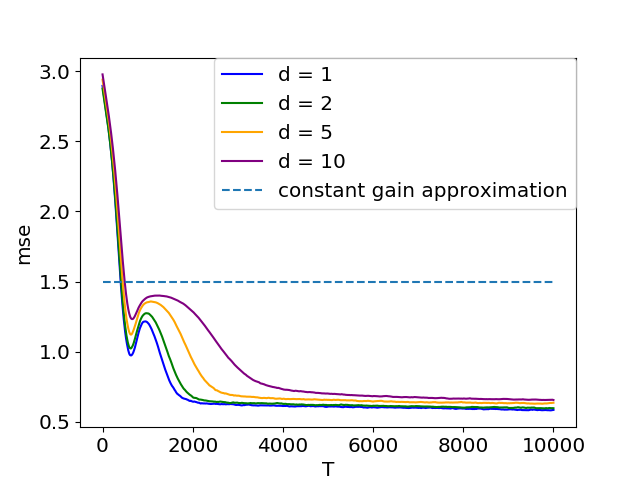}
		\caption{m.s.e as a function of iterations for varying dimensions. }
	\end{subfigure}
 &
	\begin{subfigure}[b]{0.3\linewidth}
		\includegraphics[width=\linewidth]{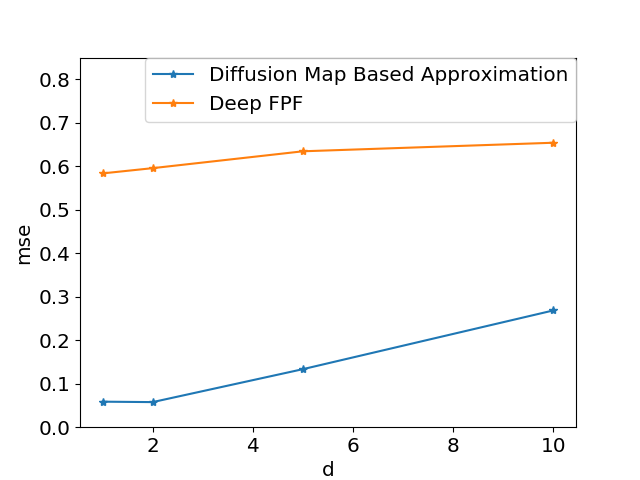}
		\caption{comparison of m.s.e as a function of dimension}
	\end{subfigure}
&
	\begin{subfigure}[b]{0.3\linewidth}
	\includegraphics[width=\linewidth]{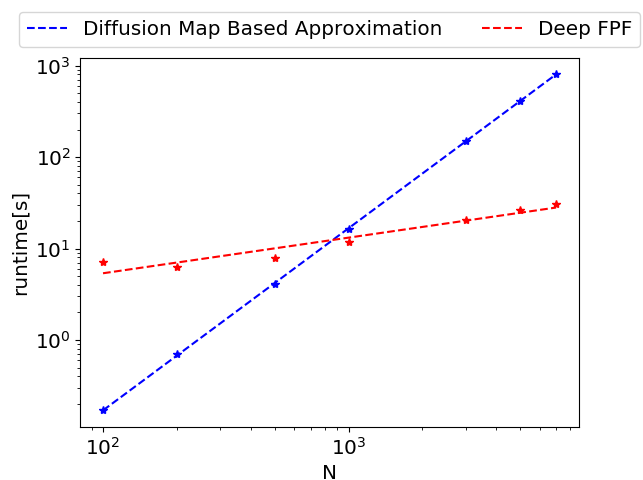}
	\caption{Runtime comparison for Deep FPF and diffusion map based approximation}
	\label{fig:deep_vs_kernel}
\end{subfigure}
\end{tabular}
	\caption{Numerical analysis of the m.s.e~\eqref{eq:mse} and the runtime for the proposed procedure and comparison with Diffusion-map (DM) and Galerkin approach, for the example in Section~\ref{subsec:scaling_with_d}.}
	\label{fig:mse_dims}
\end{figure*}

The resulting m.s.e  as a function of iteration for different dimensions is depicted in~Figure~\ref{fig:mse_dims}-(a). It is observed that the dimension does not effect the resulting error to a great degree. For comparison, the m.s.e as a function of dimension, for the proposed approach and the diffusion-map algorithm, is depicted in Figure~\ref{fig:mse_dims}-(b). It is observed that although the m.s.e for diffusion-map is smaller, but it grows faster with dimension compared to the neural- network-based approach.
Also, in our simulations we used the optimal value of the kernel-bandwidth for the diffusion-map algorithm for each dimension. In particular  $\epsilon = 0.1,0.1,0.2,1.0$  for $d=1,2,5,10$ respectively. This hyper parameter tuning may not be possible in application.

\subsection{Scaling of computational time with $N$}
Comparison of the running-time for the proposed procedure~\ref{alg:DeepGain} and Diffusion map-based approach as a function of sample size (number of particles) is depicted in Figure~\ref{fig:deep_vs_kernel}. 


It is observed that the running time of the diffusion map based approximation grows with $O(N^2)$, while the running time of neural network-based approach scales with $O(N)$. This makes the proposed approach favourable for large sample size, which is necessary for high-dimensional problems. 


\section{Application to filtering}
\label{sec:filtering_application}
Consider the problem of transporting particles  from initial distribution (or prior distribution) $\rho_0(x)$ to the final distribution (or posterior distribution) \[\rho_1(x) = \frac{\rho_0(x)e^{-l(x)}}{\int \rho_0(y)e^{-l(y)} \ud y}.\] This problem appears in filtering problem with discrete-time observations, where the function $l(x)$ represents the log-likelihood function of the observation model~\cite{daum10,daum2017generalized,yang2014continuous}. 

 The transportation is achieved by updating the particles according to
\begin{equation}\label{eq:Xi}
\frac{\ud X^i_t}{\ud t} = -\nabla  \phi(t,X^i_t),\quad X^i_0 \sim \rho_0,\quad t \in[0,1],
\end{equation}  
where $\phi(t,\cdot)$ is the solution to the Poisson equation~\eqref{eq:Poisson} with $\rho$ as the distribution of the particles $\{X^i_t\}_{i=1}^N$. 

The justification for~\eqref{eq:Xi} is as follows. Let $\rho(t,x)$ denote the distribution of the particles  $X^i_t$. We show that $\rho(1,x)$ is equal to the posterior distribution $\rho_1(x)$. The evolution of $\rho(t,x)$ is given by the continuity equation
\begin{equation*}
\frac{\partial \rho}{\partial t} (t,x) = \nabla \cdot(\rho(t,x)\nabla \phi(t,x) ) ,\quad \rho(0,x) = \rho_0(x),
\end{equation*} 
which is equal to
\begin{equation*}
\frac{\partial \rho}{\partial t} (t,x) = -\rho(t,x)(h(x)-\hat{h}_t),\quad \rho(0,x) = \rho_0(x),
\end{equation*} 
because $\phi(t,x)$ solves the Poisson equation~\eqref{eq:Poisson}. 
The solution to this pde is 
\begin{equation}\label{eq:filt_exact}
\rho  (t,x) = \frac{\rho_0(x)e^{-th(x)}}{\int \rho_0(y)e^{-th(y)} \ud y},
\end{equation} 
concluding $\rho(1,x)=\rho_1(x)$. The trajectory $\rho(t,x)$ is known as the homotopy between $\rho_0$ and $\rho_1$. 

\begin{figure*}[t]
	\centering
	\begin{tabular}{ccc}
	\begin{subfigure}[b]{0.42\linewidth}
		\includegraphics[width=0.9\linewidth]{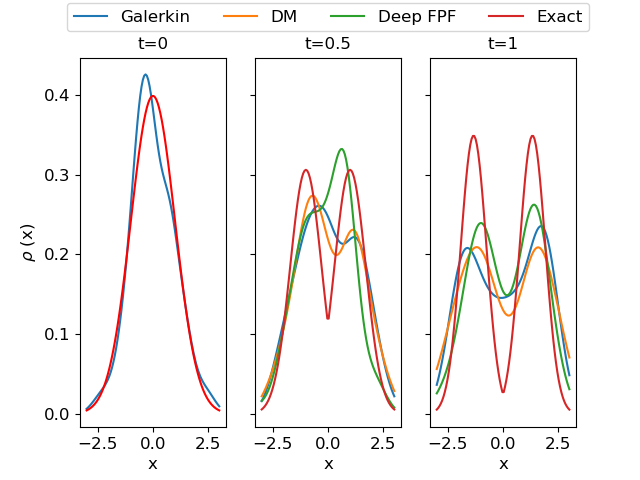}
	\caption{}
	\end{subfigure}
	&
	\begin{subfigure}[b]{0.42\linewidth}
		\includegraphics[width=0.9\linewidth]{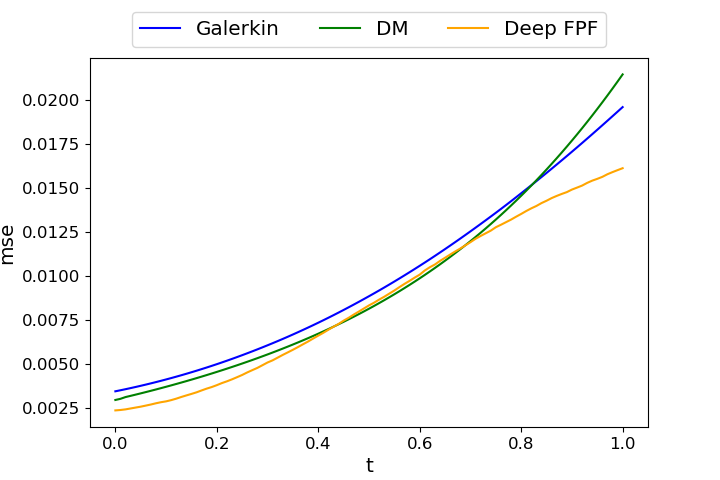}
	\caption{}
	\end{subfigure}
\end{tabular}
	\caption{The empirical distribution of the particles simulated according to~\eqref{eq:Xi} where $\nabla \phi$ is approximated using three different algorithms: Neural network-based, Diffusion map-based (DM), and Galerkin. (a) shows a kernel density estimate of the empirical distribution along with the exact distribution given by the homotopy~\eqref{eq:filt_exact}. (b) shows m.s.e in estimating the conditional expectation of $\psi(x) = {1}_{x>0}$ computed according to~\eqref{eq:mse-psi} for the filtering example of Section~\ref{sec:filtering_application}.}
	\label{fig:priori_posteriori}
\end{figure*}

For example, let $\rho_0$ be a Gaussian distribution $N(0,1)$ and let $l(x)= (|x|-2)^2$. This likelihood model induces a bimodal posterior distribution.  The resulting flow of particles, with $\nabla \phi(t,x)$ computed according to the proposed algorithm~\ref{alg:DeepGain}, the diffusion map-based algorithm with $\epsilon=0.1$, and the Galerkin algorithm with  fifth order polynomial basis functions, is depicted in Figure~\ref{fig:priori_posteriori}(a). The figure shows a kernel-density estimate of the empirical distribution of the particles along with the exact distribution $\rho(t,x)$  at three time instants: $t=0,0.5,1.0$. 

A quantitative comparison is provided by calculating the mean square error in estimating the conditional expectation of the function $\psi(x) = x {1}_{x>0}$ over time. The m.s.e is defined according to
\begin{equation}\label{eq:mse-psi}
{mse}_t = \frac{1}{K} \sum_{k=1}^K (\frac{1}{N} \sum_{i=1}^N\psi(X_t^{k,i})-\int \psi(x)\rho(t,x) dx)^2,
\end{equation}
where $K=100$ is the number of independent simulations. The result is depicted in Figure~\ref{fig:priori_posteriori}(b).

\begin{remark}
In a filtering application, it is not necessary to reinitialize the neural network for gain function approximation after each time the particles are moved. 
 Because the particles move slightly at each time step, the gain function does not vary much. Therefore, the gain function that is obtained in the previous filtering step is a good initialization. This reduces the required number of iterations for gain function approximation significantly.  
\end{remark}

\section{Conclusion}
We presented a deep learning-based approach to approximate the gain function in feedback particle filter, and provided preliminary numerical results that serves as proof of concept. There are two main directions of future work: (i) sample complexity analysis of the proposed procedure in terms of neural network architecture. This requires non-trivial application of the existing {\it generalization theory} results, because the objective function involves gradient of the neural net evaluated on samples which is not standard; (ii) numerical analysis of the proposed procedure for high-dimensional filtering problems and comparison with importance sampling-based particle filters. 

\label{sec:conclusion}

\bibliographystyle{IEEEtran}
\bibliography{refs,SIAM-Gain,TAC-OPT-FPF}

\end{document}

%% file: _definitions.tex
\def\qed{\hspace*{\fill}~\IEEEQED\par\endtrivlist\unskip}

\def\Re{\mathbb{R}}

\def\notes#1{\marginpar{\tiny #1}\typeout{Notes!
Notes!
Notes!
}}
\renewcommand{\notes}[1]{\typeout{notes!}}
\def\FRAC#1#2#3{\genfrac{}{}{}{#1}{#2}{#3}}
\def\half{{\mathchoice{\FRAC{1}{1}{2}}%
{\FRAC{2}{1}{2}}%
{\FRAC{3}{1}{2}}%
{\FRAC{4}{1}{2}}}}

\def\Re{\field{R}}

\def\clZ{{\cal Z}}

\def\Expect{{\sf E}}

\def\Expect{{\sf E}}

\def\IEEEQEDclosed{\mbox{\rule[0pt]{1.3ex}{1.3ex}}}
\def\qed{\nobreak\hfill\IEEEQEDclosed}

\def\clZ{{\cal Z}}

\newtheorem{theorem}{Theorem}

\newtheorem{remark}{Remark}
\newtheorem{proposition}{Proposition}

\def\beq{\begin{eqnarray}} 
\def\bc{\begin{center}} 
\def\be{\begin{enumerate}}
\def\bi{\begin{itemize}} 
\def\bs{\begin{small}}
\def\bS{\begin{slide}}
\def\ec{\end{center}} 
\def\ee{\end{enumerate}}
\def\ei{\end{itemize}}
\def\es{\end{small}}
\def\eS{\end{slide}}
\def\eeq{\end{eqnarray}}

\newcommand{\newP}[1]{\medskip\noindent{\bf #1:}}

\newcommand{\ud}{\,\mathrm{d}}

\def\Re{\mathbb{R}}

\def\Expect{{\sf E}}

\def\clZ{{\cal Z}}

\renewcommand{\Re}{\mathbb{R}}

\def\FRAC#1#2#3{\genfrac{}{}{}{#1}{#2}{#3}}

